\documentclass[letterpaper]{article} 
\usepackage{aaai25}  
\usepackage{times}  
\usepackage{helvet}  
\usepackage{courier}  
\usepackage[hyphens]{url}  
\usepackage{graphicx} 
\usepackage{natbib}  
\usepackage{caption} 
\usepackage[utf8]{inputenc} 
\usepackage[T1]{fontenc}    
\usepackage{url}            
\usepackage{booktabs}       
\usepackage{amsfonts}       
\usepackage{nicefrac}       
\usepackage{microtype}      
\usepackage{xcolor}         
\usepackage{algorithm}
\usepackage{algorithmic}
\usepackage{graphicx}
\usepackage{algorithm}
\usepackage{algorithmic}
\usepackage{amsmath}
\usepackage{multicol}
\usepackage{multirow}
\usepackage{amsmath}
\usepackage{newfloat}
\usepackage{listings}
\DeclareCaptionStyle{ruled}{labelfont=normalfont,labelsep=colon,strut=off} 
\floatstyle{ruled}
\newfloat{listing}{tb}{lst}{}
\floatname{listing}{Listing}
\nocopyright 

\title{Reflective Human-Machine Co-adaptation for \\Enhanced Text-to-Image Generation Dialogue System}
\author {
    Yuheng Feng\textsuperscript{\rm 1},
    Yangfan He\textsuperscript{\rm 2},
    Yinghui Xia\textsuperscript{\rm 3},
    Tianyu Shi\textsuperscript{\rm 4},
    Jun Wang\textsuperscript{\rm 5},
    Jinsong Yang\textsuperscript{\rm 3}\thanks{Corresponsding author},
}
\affiliations {
    \textsuperscript{\rm 1}Xidian University
    \textsuperscript{\rm 2}Xiamen University
    \textsuperscript{\rm 3}AutoAgents.ai\\
    \textsuperscript{\rm 4}University of Toronto
    \textsuperscript{\rm 5}East China Normal University\\
    yuhengfeng98@gmail.com, he000577@umn.edu, vix@autoagents.ai, \\tianyu.s@outlook.com, wongjun@gmail.com, edward.yang@autoagents.ai
}

\usepackage{bibentry}

\begin{document}

\maketitle

\begin{abstract}
Today's image generation systems are capable of producing realistic and high-quality images. 
However, user prompts often contain ambiguities, making it difficult for these systems to interpret users' potential intentions.
Consequently, machines need to interact with users multiple rounds to better understand users' intents. The unpredictable costs of using or learning image generation models through multiple feedback interactions hinder their widespread adoption and full performance potential, especially for non-expert users.
In this research, we aim to enhance the user-friendliness of our image generation system. To achieve this, we propose a reflective human-machine co-adaptation strategy, named RHM-CAS. Externally, the Agent engages in meaningful language interactions with users to reflect on and refine the generated images. Internally, the Agent tries to optimize the policy based on user preferences, ensuring that the final outcomes closely align with user preferences.
Various experiments on different tasks demonstrate the effectiveness of the proposed method.
\end{abstract}

\section{Introduction}

CGenerative artificial intelligence has demonstrated immense potential in facilitating economic development by helping optimize creative and non-creative tasks. Models such as DALL·E 2, IMAGEN, Stable Diffusion, and Muse have achieved this through their capability to produce unique, convincing, and lifelike images and artwork from textual descriptions\cite{gozalo2023chatgpt}. Despite the considerable progress achieved, there remains substantial potential for improvement, particularly in generating higher-resolution images that more accurately reflect the semantics of the input text and in designing more user-friendly interfaces\cite{frolov2021adversarial}. Many models find it hard to accurately comprehend the nuanced intentions behind human instructions, often leading to a mismatch between user expectations and model outputs.

Moreover, the impact of certain adjustments to variables on the final image output is not always straightforward, posing a significant challenge for non-expert users who haven't systematically learned prompt engineering courses. The intricacy involved in comprehending and manipulating these variables presents a substantial obstacle for individuals without a technical background. Furthermore, given the same input text, the model may still generate images with substantially different content or layouts, where aspects such as background, color, and perspective can vary. In such instances, the user must engage in multiple trials, and acquiring an image that meets their specific requirements can depend significantly on chance.

To address these challenges, we introduce an innovative dialogic approach designed to enhance the user experience for non-professional users. Within this dialogic interaction process, we posit the existence of a latent generative objective in the user's mind. A single image may represent the user's latent and unconscious generative goal. By iteratively querying the user, we can progressively elicit more detailed descriptions, with the ultimate aim of producing an image that closely aligns with the user's underlying intent. Figure \ref{Diagram} illustrates the operational flow of this project as interacted by the users. This approach is inspired by the concept of human-in-loop co-adaptation~\cite{reddy2022first}, where the model evolves alongside user feedback to better align with user expectations. Our main contributions are:
\begin{itemize}
\item We delve into human-machine interaction methods within image generation tasks, guiding users through the process to effectively create images that reflect their intentions and preferences.

\item We introduce an enhanced Text-to-Image dialogue based  Agent, which leverages both external interactions with users and internal reflections to enhance its performance.

\item Application across general image and fashion image generation demonstrates the versatility and potential value of our approach.
\end{itemize}

\section{Related work}
\label{sec:formatting}
\subsection*{Text-Driven Image Editing Framework}
Recent advancements in text-to-image generation have focused on aligning models with human preferences, using feedback to refine image generation. Studies range from Hertz et al.~\cite{hertz2022prompt}'s framework, which leverages diffusion models' cross-attention layers for high-quality, prompt-driven image modifications, to innovative methods like ImageReward~\cite{xu2024imagereward}, which develops a reward model based on human preferences. These approaches collect rich human feedback~\cite{wu2023better,liang2023rich}, from detailed actionable insights to preference-driven data, training models for better image-text alignment and adaptability~\cite{lee2023aligning} to diverse preferences, marking significant progress in personalized image creation.

\subsection*{Ambiguity Resolution in Text-to-Image Generation}
From visual annotations~\cite{endo2023masked} and model evaluation benchmarks~\cite{lee2024holistic} to auto-regressive models~\cite{yu2022scaling} for rich visuals, along with frameworks for abstract~\cite{liao2023text} and inclusive imagery~\cite{zhang2023iti}, the text-to-image field is advancing through strategies like masked transformers~\cite{chang2023muse}, layout guidance~\cite{qu2023layoutllm} without human input, and feedback mechanisms~\cite{liang2023rich} for quality. The TIED framework and TAB dataset~\cite{mehrabi2023resolving} notably enhance prompt clarity through user interaction, improving image alignment with user intentions, thereby boosting precision and creativity. 
\subsection*{Human Preference-Driven Optimization for Text-to-Image Generation Models}
Zhong et al.~\cite{zhong2024panacea} significantly  advance the adaptability of LLMs to human preferences with their innovative contributions. Zhong et al.'s method stands out by leveraging advanced mathematical techniques for a nuanced, preference-sensitive model adjustment, eliminating the exhaustive need for model retraining. Xu et al.~\cite{xu2024imagereward} take a unique approach by harnessing vast amounts of expert insights to sculpt their ImageReward system, setting a new benchmark in the creation of images that resonate more deeply with human desires. Together, these advancements mark a pivotal shift towards more intuitive, user-centric LLMs technologies, heralding a future where AI seamlessly aligns with the complex mosaic of individual human expectations.\\

\section{Proposed method}

\begin{figure*}[htbp]
  \centering
  \includegraphics[width=2\columnwidth]{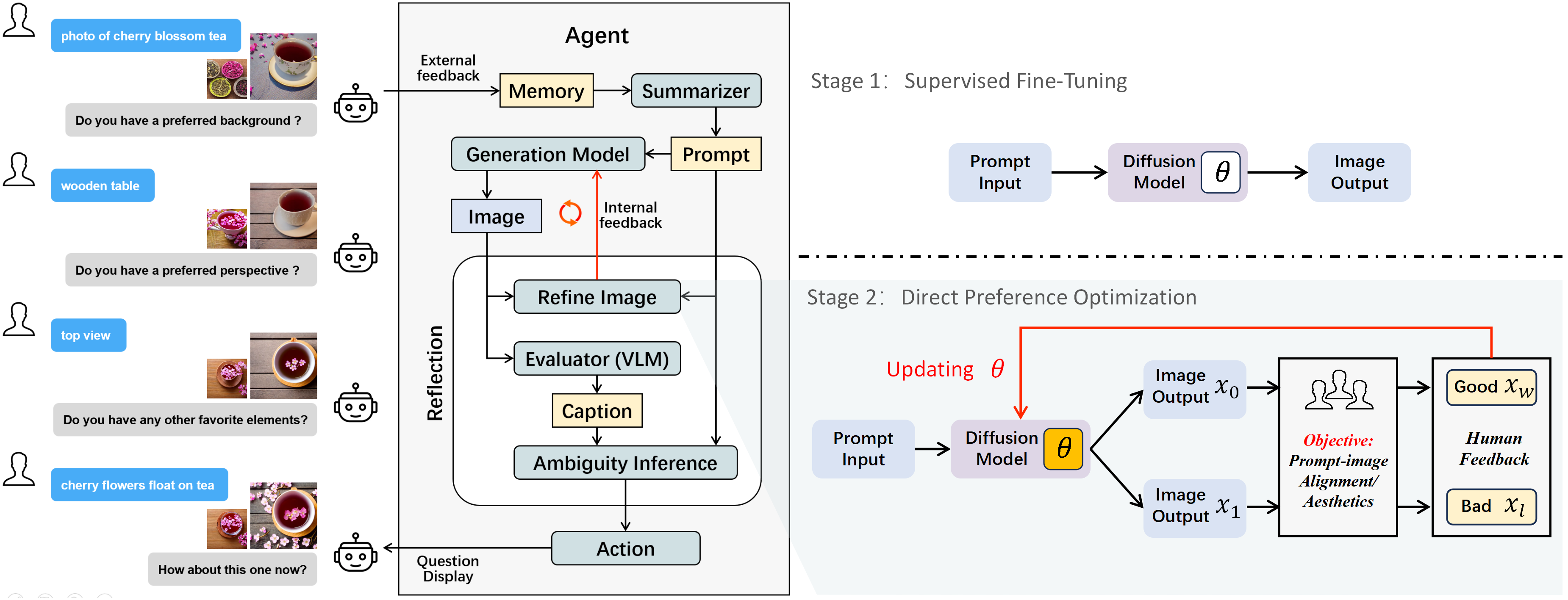} 
  \caption{Proposed framework of Enhanced Text-to-Image Reflexion Agent. The Generation Model can learn user preferences by Direct Preference Optimization.}
  \label{Diagram}
\end{figure*}

We developed a modular architecture tailored for image generation tasks within multi-turn dialogues. This architecture is designed to facilitate deep introspection of the generation system and effectively guide user interactions. The system comprises several key components: The \textit{Memory} stores the dialogue, denoted as $h$. The \textit{Summarizer}, denoted as $M_{S}$, integrates users' historical dialogue content, and generates a \textit{Prompt}, denoted as $P$, for image generation. The \textit{Generation Model}, denoted as $M_{G}$, is responsible for transforming $P$ into specific images. The \textit{Reflection Block}, denoted as $B_{R}$, plays a crucial role. It not only handles the reasoning process (completing tasks in collaboration with the user) but also engages in internal reflection on the model. Within this module, the \textit{Evaluator}, marked as $M_{E}$, is tasked with providing a comprehensive description of the generated images. The \textit{Ambiguity Inference} $M_{inf}$ analyses the potential ambiguity and outputs an internal label $r$. Finally, the \textit{Action}, designated as $M_{A}$, displays the image and poses questions to the user. We provide a detailed exposition of this interactive framework, distinguishing between its internal and external workflows.

\subsection{External Reflection via Verbal Reflection}
The external reflection is contingent on user interactions. When the user presents a new prompt, the agent generates a corresponding image and subsequently reflects on which intents to inquire about based on that image. This interactive process is termed Human-Machine Reflection (HM-Reflection).
\par
\textbf{Memory and Summarizer} \;
The historical dialogues between the user and the agent are stored in the \textit{Memory}, while the \textit{Summarizer} $M_{S}$ generates the prompt for controlling image generation based on these historical dialogues. 
Let $h$ represent the historical dialogues, $t$ represent the current time, $w_{t}$ represent the current user's response, and $P_{t}$ represent the internal prompt used for image generation. The entire process can be expressed with the following formula:
\begin{equation}
P_{t} = M_{S}(w_{t},h).
\end{equation}
\par
\textbf{Generation Model} \;
The \textit{Generation Model} $M_{G}$ is central to the image generation, creating images based on provided prompts. Besides generating images that align with user intentions, it also incorporates additional details not explicitly mentioned by the user. For the general image generation task, we use the Stable Diffusion model v1.4 \cite{Rombach_2022_CVPR}. Specifically, for the fashion image generation task, we employ a Stable Diffusion XL v1.0 \cite{podell2023sdxl}, fine-tuned on fashion-related datasets. This is because fashion images are generally uniform in layout and demand a richer representation of fine-grained features. Let $I_{t}$ represent the currently generated image. This process can be expressed as:
\begin{equation}
I_{t} = M_{G}(P_{t}).
\end{equation}
\par
\textbf{Evaluator} \;
In this interactive reflection framework, the \textit{Evaluator} $M_{E}$ plays a critical role in assessing the quality of the generated images. The \textit{Evaluator} uses a visual language model (VLM) to describe the image content and generates captions that include aspects such as content, style, and background. We utilize Qwen-VL (7B) \cite{Qwen-VL} in the general image generation task and ChatGPT 4.0 \cite{OpenAI2023ChatGPT} in the fashion image creation task, as the VLM evaluator.
The generated captions are represented as $C_{t}$, where $C_{t}$ encompasses $N$ aspects of the description. 
\begin{equation}
C_{t} = M_{E}(I_{t}),\; C_t=\left\{C_{t}^{1}, C_{t}^{2}, \ldots, C_{t}^{N}\right\}.
\end{equation}
\par
\textbf{Inference and Action} \;
By comparing the similarity between multiple captions $C_{t}$ and the prompt $P_{t}$, the \textit{Ambiguity Inference} Model $M_{inf}$ identifies which contents are expected by the user and which are randomly generated, and output an Ambiguitiy label $r_{t}$. Based on the detected ambiguities $r_{t}$, the \textit{Action} $M_{A}$ asks the user for more detailed information. Question $q_{t+1}$ can be selected from a predefined list of questions or generated by a large language model (LLM) based on the captions and prompts.
\begin{equation}
r_{t} = M_{inf}(C_{t}, P_{t}),
\end{equation}
\begin{equation}
q_{t+1} = M_{A}(C_{t}, r_{t}).
\end{equation}
\par
The entire process of external reflection has been formalized into Algorithm \ref{External}.

\begin{algorithm}[htbp]
\caption{External reflection via Verbal Reflection}
\label{External}
\begin{algorithmic}[1]
\STATE Initialize Agent: $M_{S}$, $M_{G}$, $M_{E}$, $B_{R}$, $M_{A}$
\WHILE{dialog}
    \STATE User input words: $w_{t}$
    \STATE Store $w_{t}$ into Memory $h$
    \STATE Summarizer $M_{S}$ generates Prompt $P_{t}$
    \STATE Generation Model $M_{G}$ generates Image $I_{t}$
    \STATE Reflection $B_{R}$:
    \STATE \hspace{\algorithmicindent} Evaluator $M_{E}$ generates Caption $C_{t}$
    \STATE \hspace{\algorithmicindent} Inference Ambiguity $r_{t}$
    \STATE Action $M_{A}$ generates Question $q_{t+1}$
    \STATE Store $q_{t+1}$ into Memory $h$
\ENDWHILE
\end{algorithmic}
\end{algorithm}

\subsection{Internal Reflection via Direct Preference Optimization}
An efficient intelligent interaction system not only provides effective feedback and guidance to users but also has the ability to self-reflect. As illustrated in Figure \ref{Diagram}, the Agent features a \textit{'Refine Image'} step that optimizes the model or output results. After generating multiple images, users can mark the ones they prefer. The Agent then learns user preferences from this feedback to produce images that better align with user preferences. We employ a reinforcement learning method D3PO \cite{yang2023using} for preference learning, which directly learns from user feedback without the need for training a reward model. This functionality is designated as Tool 1. Additionally, we offer Tool 2, which checks the quality of generated images and regenerates those that do not align with the corresponding prompt.
\par

\textbf{Tool 1: Direct Preference Optimization (DPO)}\; 
Figure \ref{Diagram} illustrates the method of internal reflection via DPO. In Stage 1, the generation model undergoes supervised fine-tuning to adapt to a specific generation task. In Stage 2, a certain amount of preference feedback is accumulated through multiple interactions with the user. This feedback is then used to optimize the model, resulting in more personalized outputs. 
The optimization method employed is D3PO \cite{yang2023using}, which expands the theoretical DPO into a multi-step MDP (Markov Decision Process) and applies it to diffusion models. 
\par
Given two image samples, the user selects the image they prefer, denoted as $x_{w}$, while the other sample can be represented as $x_{l}$. Using the same weight, initialize a reference model $\pi_{ref}$, and a target model $\pi_\theta$. During the denoising process, the diffusion model takes a latent $s$ as input and outputs a latent $a$. Based on the probability of $\pi_{ref}$, the overall loss of the D3PO algorithm gives:

\begin{equation}
\label{gradient}
\begin{aligned}
\mathcal{L}(\theta) = & -\mathbb{E}\left[\log \rho\left(\beta \log \frac{\pi_\theta(a^w \mid s^w)}{\pi_{\text{ref}}(a^w \mid s^w)} \right. \right. \\
& \left. \left. - \beta \log \frac{\pi_\theta(a^l \mid s^l)}{\pi_{\text{ref}}(a^l \mid s^l)}\right)\right]
\end{aligned}
\end{equation}

Here, $\beta$ is the temperature parameter that controls the deviation of $\pi_{\theta}(a|s)$ and $\pi_{ref}(a|s)$.

\begin{algorithm}[htbp]
\caption{Tool 1: Direct Preference Optimization with D3PO}
\label{DPO}
\begin{algorithmic}[1]
\REQUIRE preferred samples and the other: $x_{w}$, $x_{l}$ and Corresponding Latent: $s_{w}$, $s_{l}$, $a_{w}$, $a_{l}$; number of training epochs $N$; number of prompts per epoch $K$
\STATE Copy a pre-trained diffusion model $\pi_{ref} = \pi_\theta$. Set $\pi_{ref}$ with \texttt{requires\_grad} to \texttt{False}.
\FOR{$n = 1$ to $N$}
    \STATE Training:
    \FOR{$k = 1$ to $K$}
        
        \STATE Update $\theta$ with gradient descent using Equation~\ref{gradient}
    \ENDFOR
\ENDFOR
\end{algorithmic}
\end{algorithm}

\textbf{Tool 2: Attend-and-Excite}\; Publicly available Stable Diffusion model The publicly available Stable Diffusion model exhibits issues with \textit{catastrophic neglect}, where the model fails to generate the subjects or attributes from the input prompt. To address this issue in diffusion models and improve text-image alignment, we utilize the A\&E algorithm \cite{chefer2023attend}.
\par
First, we calculate the CLIP similarity score $Sim$ between the image and prompt. Then, we identify the neglected words by backpropagating the loss function $l=1-Sim$. During the process of regenerating the image, we use the A\&E method to activate these neglected words. Repeat the above process a certain number of times. This Tool is detailed in Algorithm \ref{AE}.

\begin{algorithm}[htbp]
\caption{Tool 2: Attend-and-Excite}
\label{AE}
\begin{algorithmic}[1]
\REQUIRE Image $I_{t}$, Prompt $P_{t}$.
\STATE Initialize $token\_list \gets empty$, Iteration Number $N$, Threshold $k$
\FOR{$n = 1$ to $N$}
    \STATE Computing the Similarity of $I_{t}$ and $P_{t}$: $Sim \gets \text{CLIP}(I_{t}, P_{t})$
    \IF{Image is OK: $Sim > k$ } 
    \STATE \textbf{break}
    \ENDIF
    \STATE Computing the Objective: $l \gets 1 - Sim$
    \STATE Computing  $P_{t}$ gradient by $l$: $\Delta P_{t}$
    \STATE Locate peak value of $\Delta P_{t}$ to get $token\_id$ 
    \STATE Append $token\_id$ to $token\_list$
    \STATE Regenerate $I_{t}$ by $\textbf{A\&E}(P_{t}, token\_list)$
\ENDFOR
\RETURN Image $I_{t}$
\end{algorithmic}
\end{algorithm}

\begin{figure*}[htbp]
  \centering
  \includegraphics[width=1.6\columnwidth]{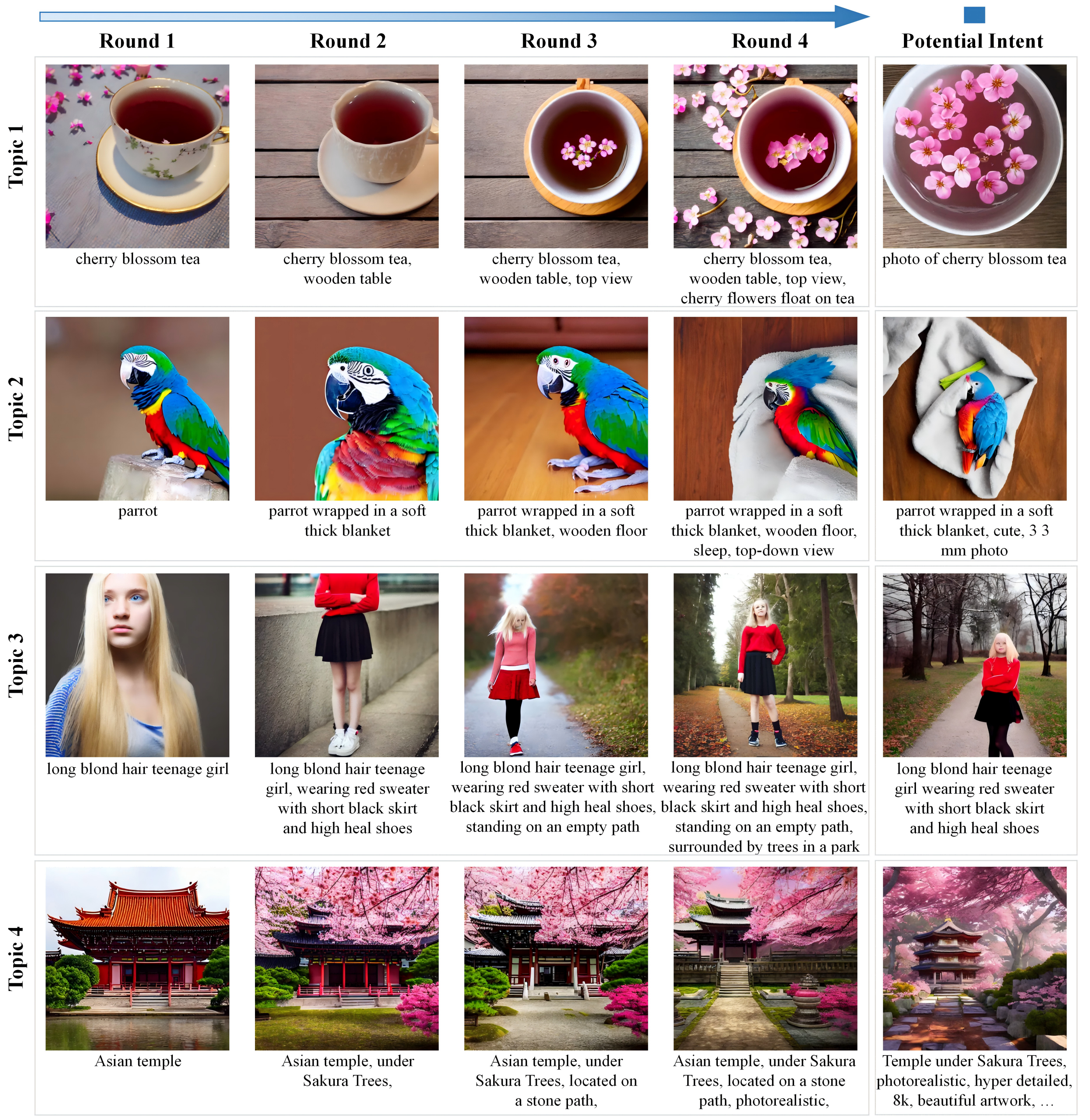} 
  \caption{ A comparative display of four rounds of image generation based on specific prompts, including cherry blossom tea, a parrot, a teenage girl, and an Asian temple across different rounds. }
  \label{task1}
\end{figure*}

\begin{table*}[htbp]
\centering
\renewcommand{\arraystretch}{1.5} 
\caption{Evaluations of prompt-intent alignment, image-intent alignment and human voting across various methodologies and integrations. Augmentation refers to using LLMs to infer ambiguity and enhance the initial prompt. HM-Reflection is the external reflection of our RHM-CAS.}
\label{baseline}
\resizebox{\textwidth}{!}{
\begin{tabular}{lccccc}
\toprule
\multirow{2}{*}{\textbf{Methods}} & \multicolumn{2}{c}{\textbf{Prompt-Intent Alignment}} & \multicolumn{2}{c}{\textbf{Image-Intent Alignment}} &   \multirow{2}{*}{\textbf{Human Voting}} \\
\cline{2-3} \cline{4-5}
                     & \textbf{T2I CLIPscore}& \textbf{T2I BLIPscore}&\textbf{I2I CLIPscore}& \textbf{I2I BLIPscore}&   \\
\midrule
GPT-3.5 augmentation & 0.157  & 0.145 & 0.624 & 0.633 & 4\% \\
GPT-4 augmentation   & 0.163  & 0.152 & 0.648 & 0.637 & 3.2\% \\
LLaMA-2 augmentation & 0.112  & 0.132 & 0.593 & 0.571 & 6\% \\
Yi-34B augmentation  & 0.101  & 0.123 & 0.584 & 0.560 & 4.4\% \\
\midrule
HM-Reflection        & 0.282  & 0.281 & 0.752 & 0.760 & 25.5\% \\
HM-Reflection + ImageReward RL  & 0.292  & 0.283 & 0.782 & 0.776 & 26.2\% \\
RHM-CAS (Ours)       & \textbf{0.328}  & \textbf{0.334} & \textbf{0.802} & \textbf{0.813} & \textbf{30.6\%} \\

\bottomrule
\end{tabular}
}
\end{table*}

\begin{table*}[htbp]
\centering
\renewcommand{\arraystretch}{1.5}
\caption{Multi-dialog (HM-Reflection) ablation experiment with image-to-image similarity scores across different rounds, including SD-1.4, SD-1.5, DALL-E.}
\label{HM_Ablation}
\resizebox{\textwidth}{!}{
\begin{tabular}{ccccccc}
\toprule
\multirow{2}{*}{\textbf{Multi-dialog}}  & \multicolumn{2}{c}{\textbf{SD-1.4}} & \multicolumn{2}{c}{\textbf{SD-1.5}}& \multicolumn{2}{c}{\textbf{DALL-E}}  \\
\cline{2-3} \cline{4-5} \cline{6-7}
                            &\textbf{I2I CLIPscore}& \textbf{I2I BLIPscore} &\textbf{I2I CLIPscore}& \textbf{I2I BLIPscore} &\textbf{I2I CLIPscore}& \textbf{I2I BLIPscore} \\
\midrule

Round 1 & 0.726 & 0.702 & 0.722 &0.698 & 0.650 & 0.673\\
Round 2 & 0.757 ($\uparrow$ 0.031) & 0.737 ($\uparrow$ 0.035) & 0.745 ($\uparrow$ 0.023) & 0.724 ($\uparrow$ 0.026)  & 0.673 ($\uparrow$ 0.023) & 0.689 ($\uparrow$ 0.016) \\
Round 3 & 0.775 ($\uparrow$ 0.049) & 0.762 ($\uparrow$ 0.060) & 0.772 ($\uparrow$ 0.050) & 0.783 ($\uparrow$ 0.085)  & 0.690 ($\uparrow$ 0.040) & 0.717 ($\uparrow$ 0.044) \\
Round 4 & 0.802 ($\uparrow$ 0.076) & 0.823 ($\uparrow$ 0.121) & 0.788 ($\uparrow$ 0.066) & 0.810 ($\uparrow$ 0.112)  & 0.741 ($\uparrow$ 0.091) & 0.735 ($\uparrow$ 0.062) \\
\bottomrule
\end{tabular}
}
\end{table*}

\begin{figure*}[htbp]
  \centering
  \includegraphics[width=2\columnwidth]{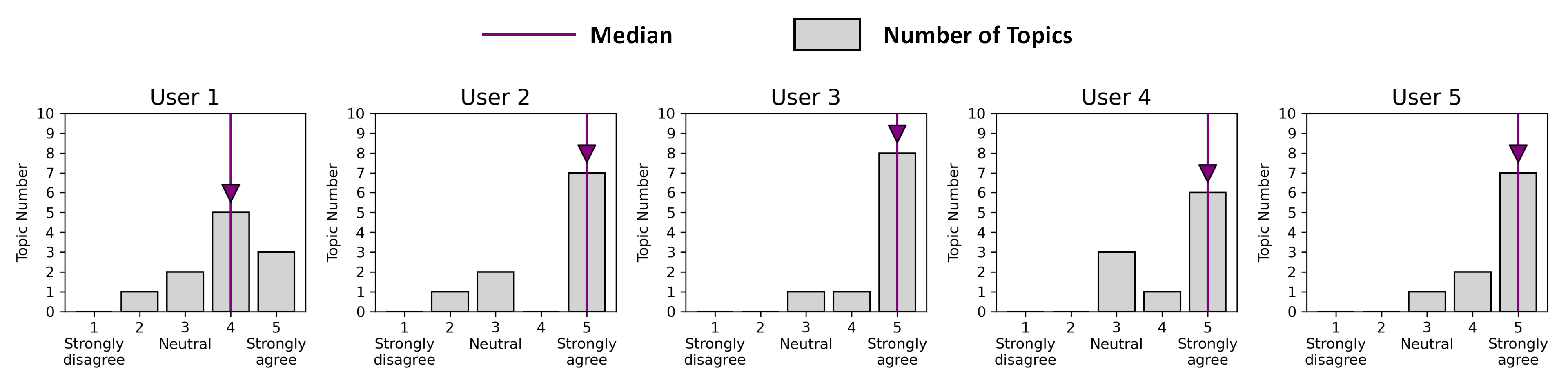} 
  \caption{Human Voting for Statement: Multi-turn dialogues can approximate the user's potential intents.}
  \label{humanVotingMulti}
\end{figure*}

\section{Experiment}
We explore the application of our proposed Enhanced Text-to-Image Reflexion Agent in two distinct scenarios: general image generation and specific fashion product creation. Due to the different requirements of these applications, adjustments have been made to our approach accordingly. In the experiments, the focus varies between the two tasks. For the general image generation task, we emphasize the effectiveness of our external reflection via verbal reflection. The emphasis of the fashion product creation task is placed on capturing fine-grained features within the images and addressing user preferences.
\par

\subsection{Task 1 General Image Generation}
The General Image Generation Task, powered by the Enhanced Text-to-Image Reflexion Agent, is designed to enhance the user experience in image creation. Our agent not only generates images based on textual instructions but also engages in dynamic dialogues with users, ensuring the images align more closely with their underlying intentions. This interactivity ensures that the images are not only visually appealing but also meet the content expectations and needs of the users. Moreover, through real-time feedback loops and continuous interaction, the agent guides users and enhances their creative expression, allowing even those with minimal experience to easily produce professional-level images.
\par

\subsubsection{Setting}
In this task, the process begins with the Summarizer generating prompts by aggregating the user's input words. These prompts are then used to generate images. The generated images are subsequently captioned by Qwen-VL \cite{Qwen-VL}, a Vision-Language Model, covering seven aspects: 'Content', 'Style', 'Background', 'Size', 'Color', 'Perspective', and 'Other'. By comparing the CLIP text similarity scores between the user's historical inputs and each caption, we identify which aspects of the image contain ambiguity. From the three aspects with the lowest scores, one is randomly selected for questioning. The question is displayed, and the user can choose whether to respond.
\par
To quantify the effectiveness of human-in-the-loop image generation, we assumed a reference image as the user's generation target in the experiments. After each image generation, the user responds based on the content of the target image until a certain number of iterations are completed. The similarity between each generated image and the target image is then evaluated to assess the effectiveness of our approach.

\subsubsection{Data Collection}
We collected those high-scoring image-text pairs from the ImageReward \cite{xu2024imagereward} dataset, which were gathered from real users. These high-scoring images exhibit excellent visual quality and a high degree of consistency with the original prompts. We excluded samples that were abstract or difficult to understand, as well as those with excessively long input prompts. Ultimately, we obtained 496 samples covering a variety of subjects, including people, animals, scenes, and artworks. And obtained over 2000 prompts from users for image generation. Some of these images also contained content not explicitly mentioned in the original prompts. These reference images served as potential targets for multi-turn dialogue generation, with each sample undergoing at least four rounds of dialogue.

\subsubsection{Baseline setup}
To demonstrate the effectiveness of our Reflective Human-Machine Co-adaptation Strategy in uncovering users' underlying intentions, we established several baselines. One approach to resolving ambiguity in user prompts is to use Large Language Models (LLMs) to rewrite the prompts. We employed several LLMs to augment the initial prompts, allowing these models to infer the users' intentions. These LLMs include: \textbf{ChatGPT-3.5}, \textbf{ChatGPT-4} \cite{achiam2023gpt}, \textbf{LLaMA-2} \cite{touvron2023llama}, and \textbf{Yi-34B} \cite{ai2024yi}. The relevant experiments are shown in Table \ref{baseline}. Table \ref{baseline} presents the alignment between the generated prompt and target image, as well as the alignment between the output image and target image. A subjective visual evaluation (Human Voting) was used to select the image result that most closely resembles the target image. All experiments were conducted on four Nvidia A6000 GPUs. The diffusion model SD-1.4 employed the DDIM sampler.
\par

Additionally, we validated the effectiveness of our Multi-dialog (HM-Reflection) approach in uncovering users' underlying intentions by using different generative models. The relevant experiments are shown in Table \ref{HM_Ablation}, including \textbf{Stable Diffusion (v1.4)}, \textbf{Stable Diffusion (v1.5) } \cite{Rombach_2022_CVPR}, and \textbf{DALL-E} \cite{ramesh2021zero}.

\subsubsection{Result Analysis}
In Figure \ref{task1}, we illustrate our reflective human-machine co-adaptation strategy. The rightmost side of the figure shows the target images observed by users during testing, serving as the users' intended generation targets. The four columns of images on the left correspond to the image results and prompt outputs at different dialogue turn. From the visual results, it is evident that by incorporating comprehensive descriptions across the seven aspects, the generated images increasingly align with the target images. 
\par
Tables \ref{baseline} and Table \ref{HM_Ablation} describe the experiments conducted on our collected dataset. Table \ref{baseline} uses the SD-1.4 as the generative model and Qwen-VL as the evaluator. It first compares the effectiveness of non-human-machine methods (LLM augmentation) in inferring user intent and then evaluates the performance of our multi-dialog approach (HM-Reflection). We compare our RHM-CAS method with a reinforcement learning approach using the feedback of ImageReward model \cite{xu2024imagereward} to improve the generative model. In Table \ref{baseline}, 'Intent' refers to the target images in the experiments. We use CLIP \cite{radford2021learning} and BLIP \cite{li2022blip} to extract embeddings of prompts and images and measure their similarity scores with the Intent embeddings. Table \ref{baseline} also includes user votes on which method produced outputs closest to the target images. Compared to other methods, our approach achieved optimal performance. Table \ref{HM_Ablation} shows the effectiveness of multi-dialog (HM-Reflection) in resolving ambiguity across different generative models. As the number of dialog rounds increases, the generated images increasingly resemble the target images, with scores in parentheses indicating the improvement relative to the initial scores.

Figure \ref{humanVotingMulti} collects the approval ratings from five testers. In these sets of dialogues conducted by each of the five users, we explore whether the users agree that the multi-round dialogue format can approximate the underlying generative target. In most cases, HM-Reflection produces results that more closely align with user intent. Besides, the experiments related to \textbf{Tool 2: Attend-and-Excite} are provided in the Appendix.

\begin{figure*}[htbp]
  \centering
  \includegraphics[width=1.35\columnwidth]{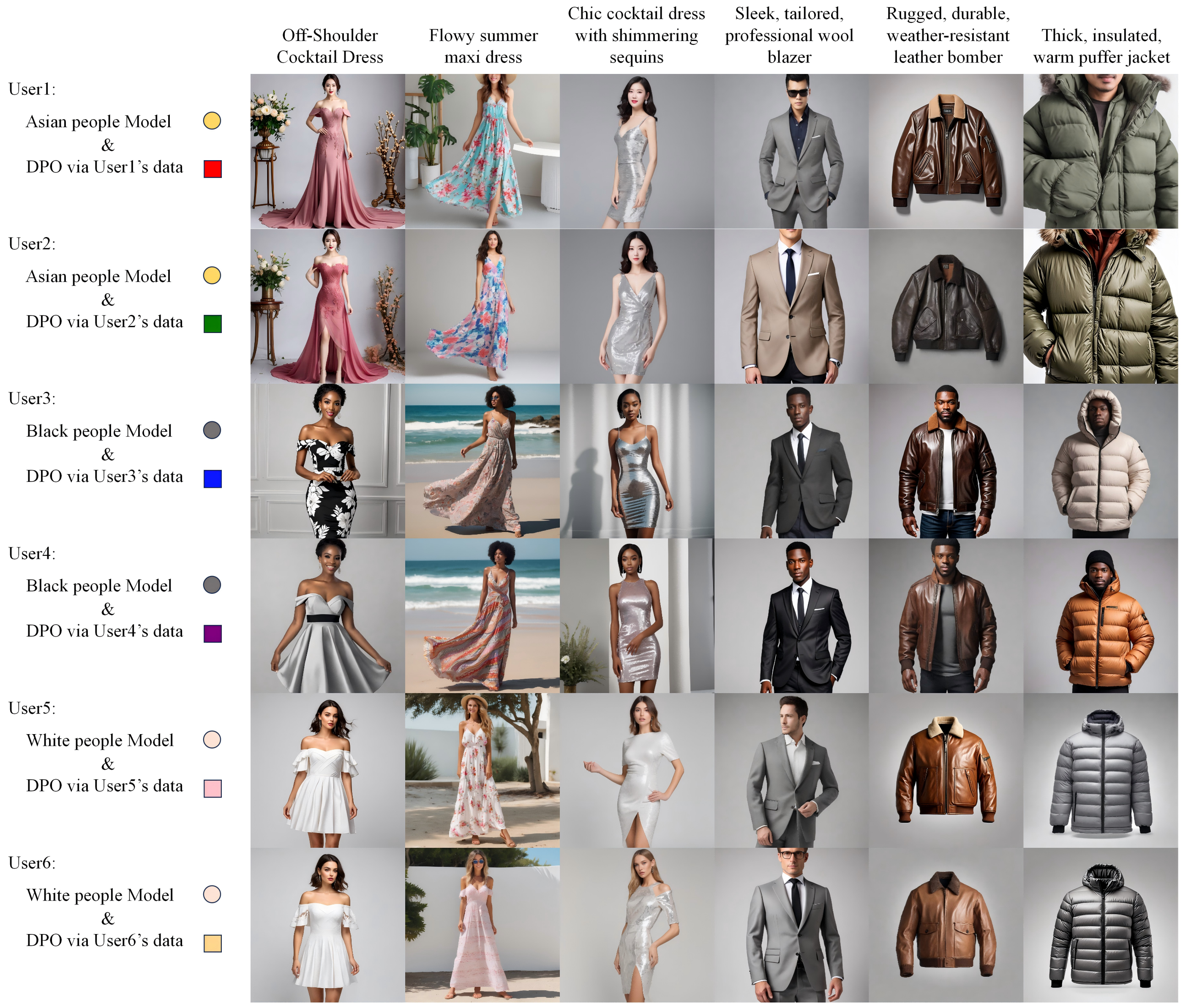} 
  \caption{ This image showcases a diverse collection of fashion models and outfits, segmented by user preferences or data. Each section highlights different styles of attire, including elegant dresses and professional to casual jackets, modeled by individuals of different ethnic backgrounds. }
  \label{task2}
\end{figure*}

\subsection{Task 2 Fashion Product Creation}
Our second task is fashion product creation, a key application of image generation technology. In the future, generating fashion products like dresses and jackets that users can purchase or customize holds great potential. This approach combines personalization and automation, offering highly customized shopping experiences. Users can generate ideal designs through simple text descriptions, reducing trial and error costs. Brands and designers can quickly test market reactions, lower inventory risks. Overall, image generation technology in fashion has a promising future.

\subsubsection{Setting}
Fashion product creation is more challenging than general image generation due to higher demands on image quality and diversity. Our Agent system also requires enhanced reasoning and multimodal understanding capabilities. During the experiments, we used ChatGPT 4.0 for reasoning tasks beyond image generation, facilitating multimodal dialogues.

For image generation, we used the SD-XL 1.0 model for its superior capabilities. We referred to the DeepFashion dataset \cite{liuLQWTcvpr16DeepFashion} for clothing types and attributes, creating labels for collecting SD-XL 1.0 image samples. These images were cleaned and curated for fine-tuning, resulting in more stable and consistent outputs. The LoRA \cite{hu2021lora} method was used for fine-tuning on four Nvidia A6000 GPUs.

To offer a customized user experience, we trained multiple models with different data, allowing users to choose models with different ethnicities. Based on user feedback, the model performs Direct Preference Optimization (DPO). In the DPO process, model parameters are updated after every 40 user feedback instances, repeated three times. The model uses the DDIM sampler for image generation.

\subsubsection{Result Analysis}
In Figure \ref{task2}, we display the outputs of six models used by different users, each optimized based on their initial model selections and interaction history. All models generated fashion products from the same prompt using identical seeds, resulting in subtle variations among the products. 

We input the same prompt into each of the six models under consistent conditions to produce six sets of fashion items. These products were then processed through Fashion-CLIP \cite{Chia2022}, a version of CLIP fine-tuned for the fashion domain, to obtain their embedding representations, which were visualized in a low-dimensional space using the t-SNE method. The visualization shows distinct preference distributions for each user in Figure \ref{task2Preference}.

Additionally, we had the six testers compare the outputs from models optimized with DPO and those without optimization. As shown in Figure \ref{humanVotingDPO}, in the majority of cases, testers believed that the DPO method improved the model's output results, more aligned with their tastes.

\section{Conclusion}
In this study, we explored the application of advanced image generation techniques integrated with human-machine interaction frameworks to enhance personalization and visual appeal in both general image generation and fashion product creation. Our Enhanced Text-to-Image Reflection System demonstrated significant capabilities in guiding users to articulate their generative intentions effectively. By leveraging both external interactions and internal reflections, our agent was able to learn from human feedback and align its outputs more closely with user preferences.
Future work will focus on integrating finer user feedback mechanisms and leveraging advancements in AI to further enhance the generative process, aiming to broaden the applicability and effectiveness of these technologies in various domains.

\section{Limitations}
This study, although advanced with the RHM-CAS, has certain limitations. In the interaction process, due to prompts containing multiple high-level descriptions, the image generation model might not fully transform all of them into images. Moreover, the VL model's ability to capture fine-grained details is limited, which may result in inaccurate captions. These cross-modal information transfer processes also lead to errors in information propagation, obstructing the expression of user intent, and thereby affecting communication efficiency. Apart from this, the method is computationally intensive, requiring substantial resources, which may limit its accessibility for users with less powerful hardware. Furthermore, the iterative refinement process, while effective, can be time-consuming. This could potentially lead to user frustration in time-sensitive situations.

Future efforts should aim to enhance computational efficiency and broaden the system's ability to generalize across more diverse inputs, improving usability in real-world applications.

\bibliographystyle{aaai25}
\bibliography{aaai25}

\appendix

\section{Q\&A Software Annotation Interface}
\label{appendix : a16}
\begin{figure}[!h]
    \centering
    \includegraphics[width=0.5\textwidth]{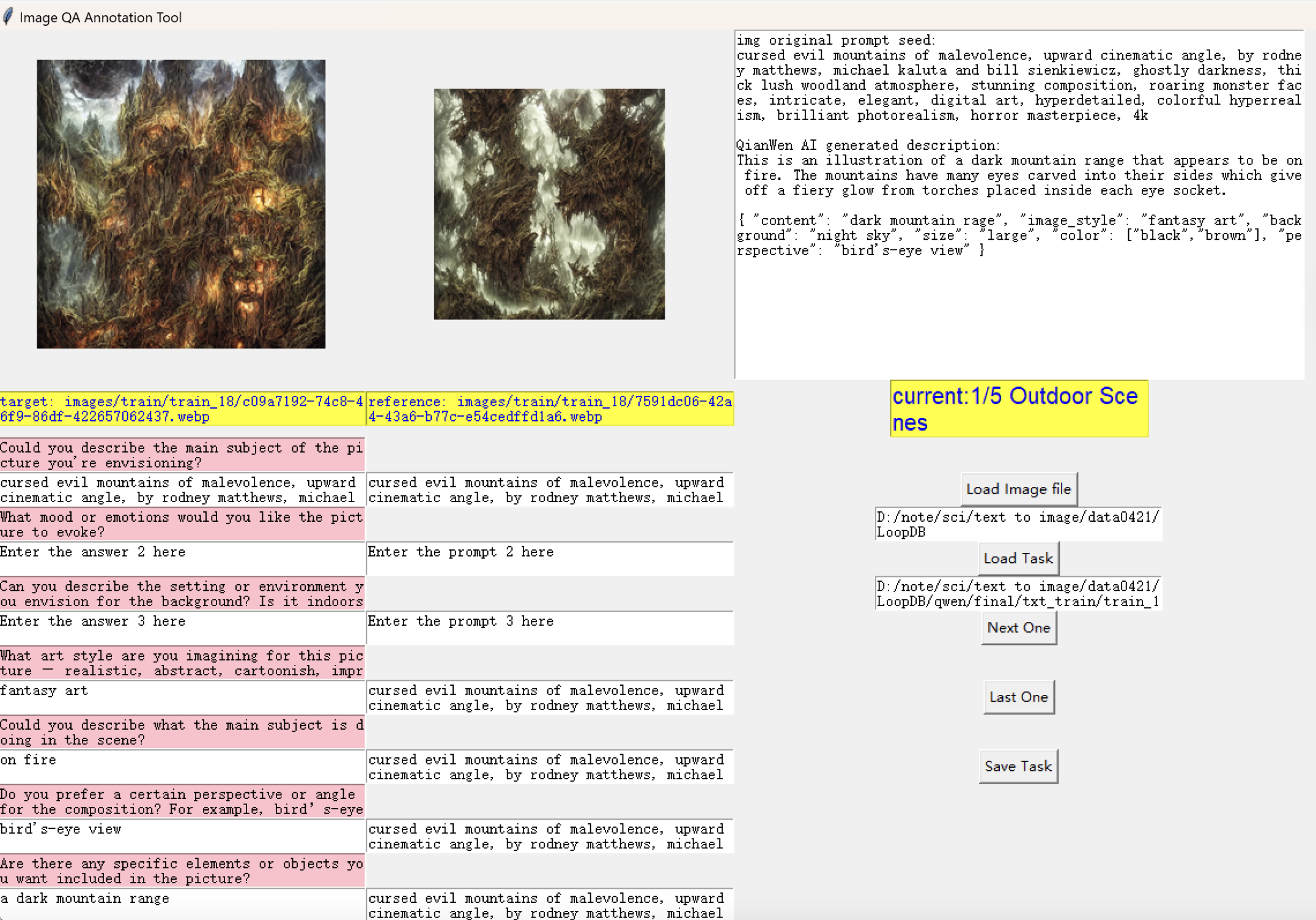} 
    \caption{Screenshot of the Q\&A software annotation interface.}
    \label{fig:qa_software_annotation_interface}
\end{figure}

Image Panel: Two images are displayed side-by-side for comparison or annotation. These images seem to depict artistic or natural scenes, suggesting the software can handle complex visual content.

HTML Code Snippet: Below the images, there's an HTML code snippet visible. This could be used to embed or manage the images within web pages or for similar digital contexts.

Interactive Command Area: On the right, there is an area with various controls and settings:

Current task and image details: Displayed at the top, indicating the task at hand might be related to outdoor scenes.
Navigation buttons: For loading new images and navigating through tasks.
Annotation tools: Options to add text, tags, or other markers to the images.
Save and manage changes: Buttons to save the current work and manage the task details.

\subsection{Human annotation instruction}

\section*{Objective}
Accurately describe and tag visual content in images to train our machine learning models.

\section*{Steps}
\begin{enumerate}
  \item \textbf{Load Image:} Use the 'Load Image' button to begin your task.
  \item \textbf{Analyze and Describe:}
  \begin{itemize}
    \item Examine each image for key features.
    \item Enter descriptions in the text box below each image.
  \end{itemize}
  \item \textbf{Tagging:}
  \begin{itemize}
    \item Apply relevant tags from the provided list to specific elements within the image.
  \end{itemize}
  \item \textbf{Save Work:} Click 'Save Task' to submit your annotations. Use 'Load Last' to review past work.
\end{enumerate}

\section*{Guidelines}
\begin{itemize}
  \item \textbf{Accuracy:} Only describe visible elements.
  \item \textbf{Consistency:} Use the same terms consistently for the same objects or features.
  \item \textbf{Clarity:} Keep descriptions clear and to the point.
\end{itemize}

\section*{Support}
For help, access the 'Help' section or contact the project manager at [contact information].

\textbf{Note:} Submissions will be checked for quality; maintain high standards to ensure data integrity.

\begin{figure*}[htbp]
  \centering  \includegraphics[width=2\columnwidth]{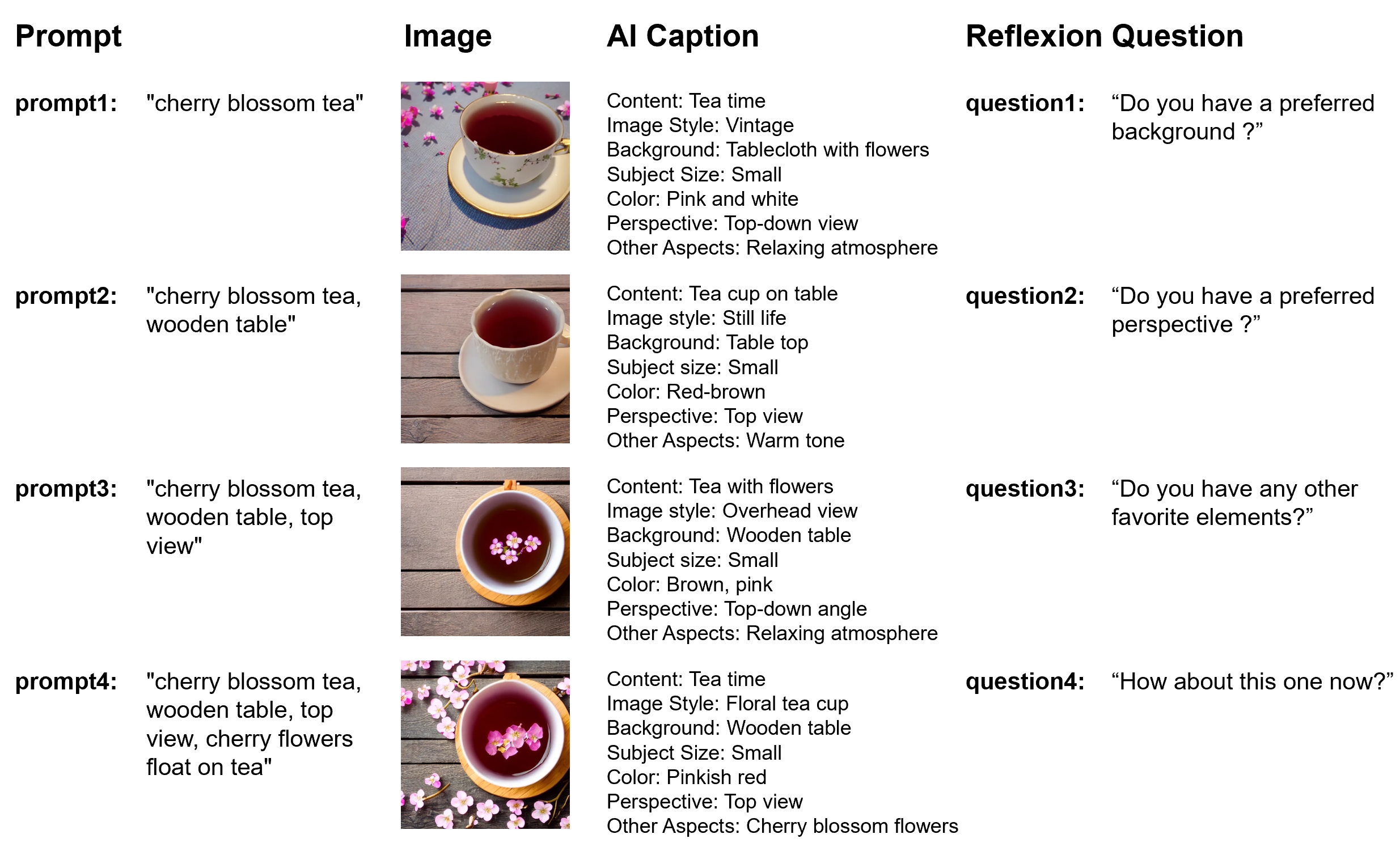} 
  \caption{ Dialogue Record of General Image Generation, including Prompts, Qwen-VL Captions and Questions. }
  \label{shortcaptions}
\end{figure*}

\section{RHM-CAS Pipeline Example}
\subsection{general image generation task pipeline}
RHM-CAS uses the Qwen-VL as the evaluator when performing general image generation tasks. Figure \ref{shortcaptions} presents an example. On the far left is the prompt generated by the Summarizer based on the user's historical dialogues, using the simplest method of phrase stacking for this task. The diffusion model then generates an image based on the current prompt. This image is subsequently described by the Qwen-VL model, which generates captions covering various aspects including "Content," "Image Style," "Background," "Subject Size," "Color," "Perspective," and "Other Aspects." The prompt and the captions are then compared, and a question related to a specific aspect is extracted from the question list.
\par

\begin{figure*}[htbp]
  \centering  \includegraphics[width=2\columnwidth]{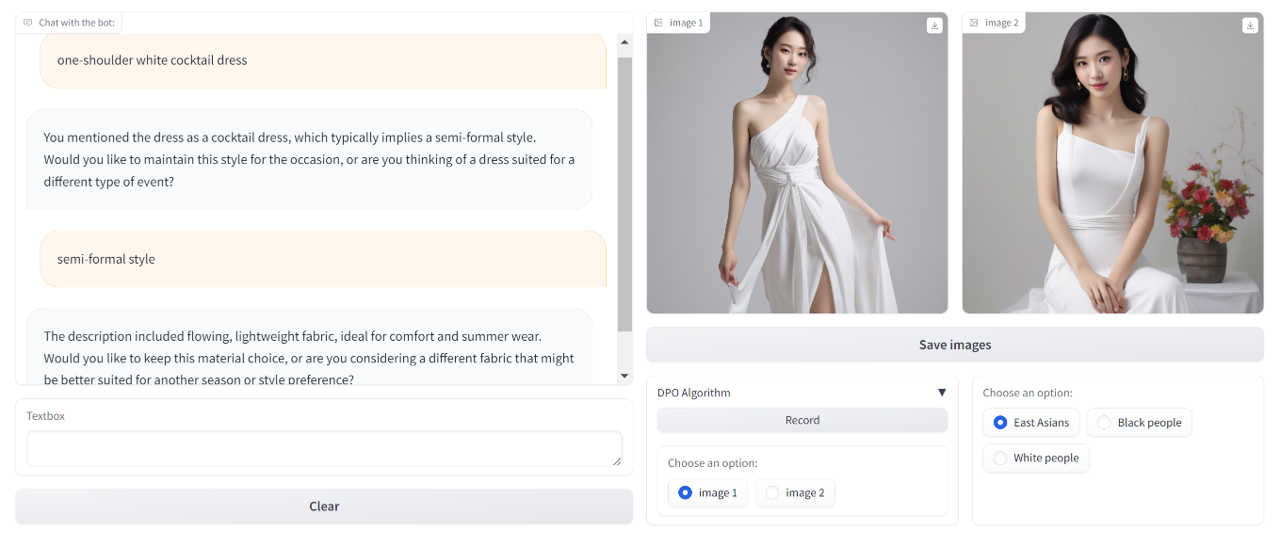} 
  \caption{Demo of Fashion Product Creation}
  \label{webdemo}
\end{figure*}

\begin{figure*}[htbp]
  \centering  \includegraphics[width=1.9\columnwidth]{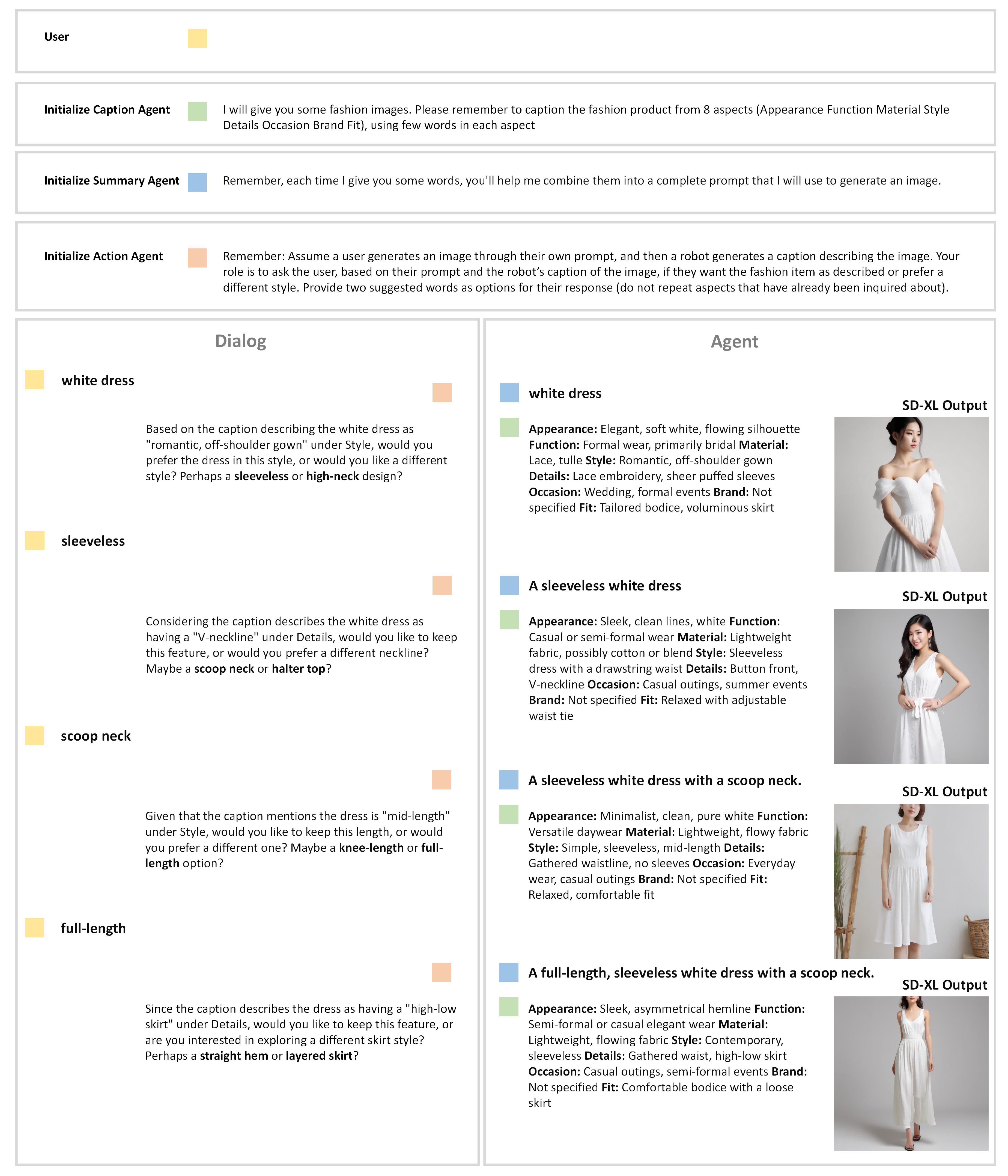} 
  \caption{Mode of Fashion Product Creation based on ChatGPT. Special Prompt initialized Each Agent.}
  \label{chatgpt}
\end{figure*}

\subsection{fashion product creation task pipeline}
When generating fashion products, we attempted to use LLMs to handle all tasks other than image generation. We selected ChatGPT-4 to manage all textual interactions with users and image descriptions, while the generative model used was our fine-tuned Stable Diffusion XL model. As shown in Figure 7, we first initialized several modules based on ChatGPT-4, including \textit{Summarizer}, \textit{Evaluator}, and \textit{Action}. Yellow represents the user's role, while other colors represent different modules of our RHM-CAS. When captioning, the Evaluator provided descriptions from multiple aspects, including 'Appearance,' 'Function,' 'Material,' 'Style,' 'Details,' 'Occasion,' and others. It can be seen that through our RHM-CAS, users can dynamically adjust the generated images and make selections based on recommendations posed by the LLM, allowing even users without prior experience to adapt quickly.
\par
Figure \ref{webdemo} showcases our demo developed based on ChatGPT. The left side of the interface is dedicated to dialogues with users, while the right side generates images in real-time based on the current conversation. The system presents two images, allowing users to choose the one they prefer, which is then used to optimize the generative model through DPO. Before using the system, users can select different ethnicities in the bottom right corner to initialize the generative model.

\section{DPO User Study}

\begin{figure*}[htbp]
  \centering
  \includegraphics[width=2\columnwidth]{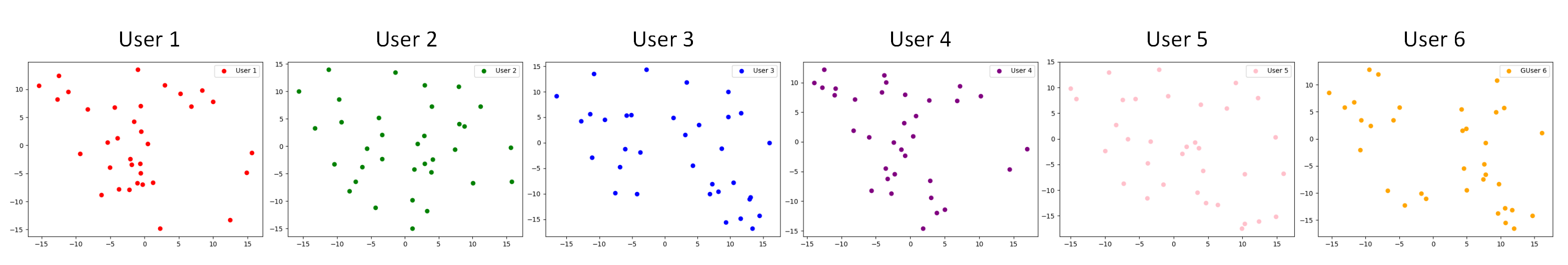} 
  \caption{Fashion-CLIP Embeddings of 6 Users visualized with t-SNE}
  \label{task2Preference}
\end{figure*}

In the fashion product creation task, we collected feedback from six users and used this feedback to optimize the model through DPO. As shown in Figure \ref{task2Preference}, under the same random seed conditions, these six models, which have been optimized multiple times, generate images using the same textual input. These images are then fed into the Fashion-CLIP \cite{Chia2022} model for embedding representation. Finally, these embedding vectors are visualized using the t-SNE method. From the latent space of Fashion-CLIP, it is evident that each of the six models exhibits distinct distribution characteristics.
\par
In addition, we invited these users to evaluate the effectiveness of DPO. Based on their assessments, in most cases, using DPO significantly improved the output performance of the model compared to the unoptimized version.

\begin{figure*}[htbp]
  \centering  \includegraphics[width=2\columnwidth]{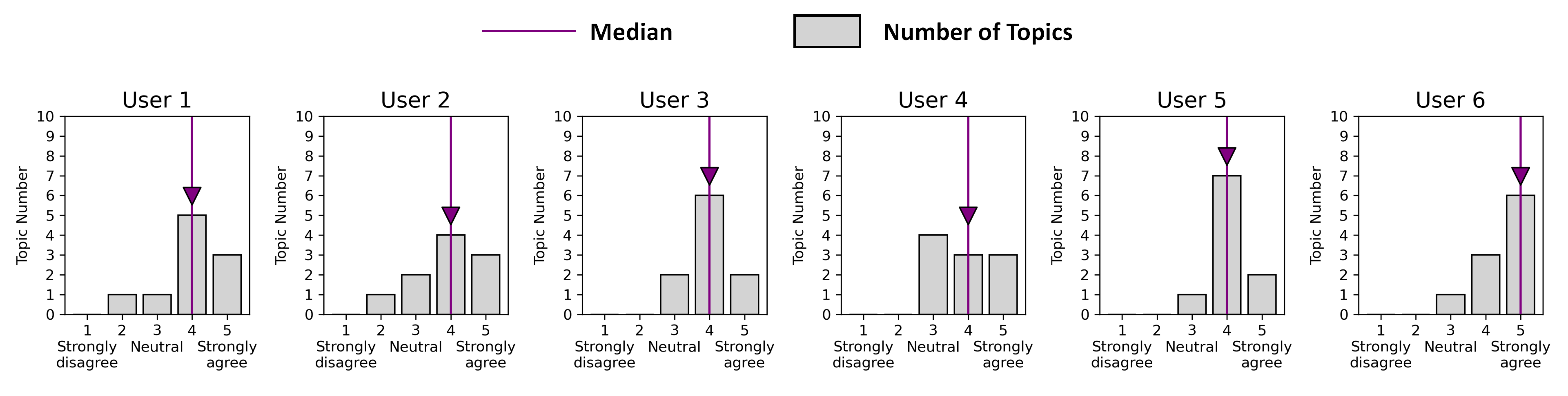} 
  \caption{Human Voting for Statement: Direct Preference Optimization can improve generation results.}
  \label{humanVotingDPO}
\end{figure*}

\begin{figure*}[htbp]
  \centering  \includegraphics[width=2\columnwidth]{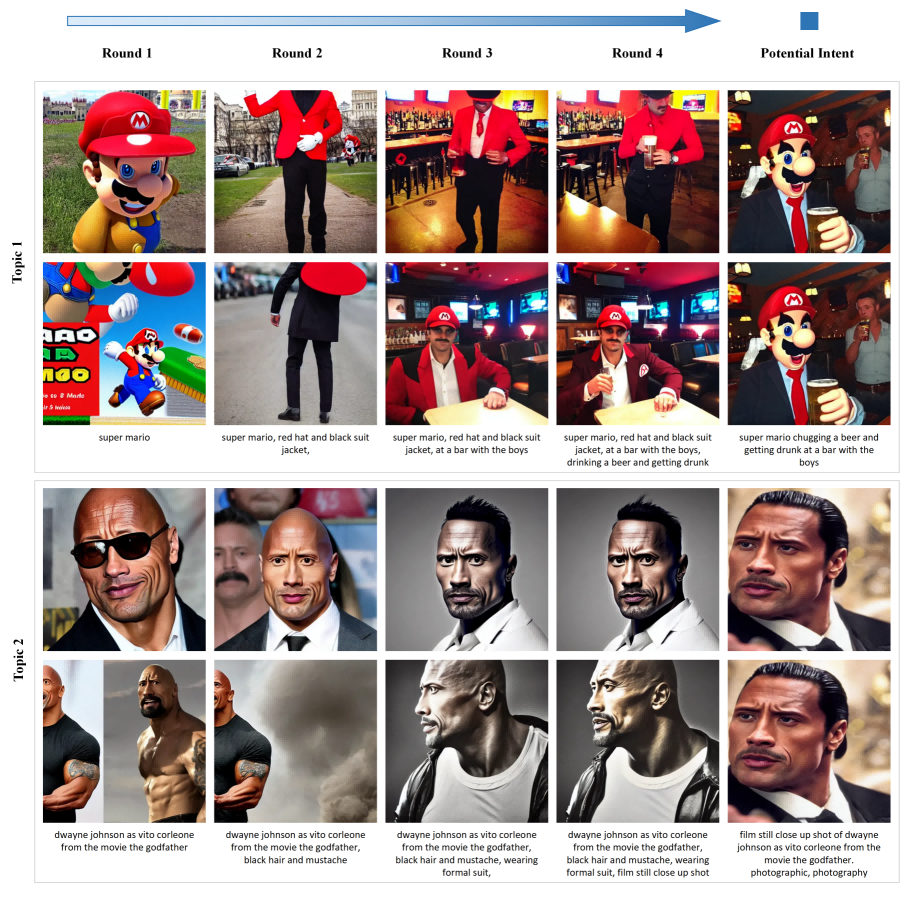} 
  \caption{Flawed Case}
  \label{failure}
\end{figure*}

\section{Tool 2 ttend-and-Excite Experimrnt}
\label{Appendix:Tool2}
We conducted independent experiments on Algorithm \ref{AE} (Tool 2: Attend-and-Excite) using the dataset collected from Task 1. As shown in Table \ref{tab:tool2_thresholds}, the second row records the usage frequency of Tool 2 as the threshold $k$ varies. When the threshold $k$ is set to 0.72 and 0.7, the usage frequencies are 31.1\% and 51.1\%, respectively. Correspondingly, the CLIP scores increased by 1.8\% and 2.3\%, indicating that these settings effectively enhance the alignment between images and text.

\begin{table*}[htbp]
\centering
\begin{tabular}{ccccccc}
\toprule
\textbf{Tool 2 threshold} & \textbf{0.8} & \textbf{0.75} & \textbf{0.72} & \textbf{0.7} & \textbf{0.68} & \textbf{0.66} \\
\midrule
\textbf{Frequency of Usage} & 0 & 8.9\% & 31.1\% & 51.1\% & 73\% & 95.5\% \\
\midrule
\textbf{T2I Similarity Improvement } & 0 & 0.2\% & 1.8\% & 2.3\% & 2.6\% & 1.0\% \\
\bottomrule
\end{tabular}
\caption{Tool 2 usage frequency and T2I Similarity at Different Tool 2 Thresholds}
\label{tab:tool2_thresholds}
\end{table*}

\section{Flawed Example}
However, we encountered some suboptimal cases during our experiments. As shown in Figure \ref{failure}, in the first topic discussing 'Super Mario', the model generated multiple rounds of images based on random noise. As the prompt length increased, the model's understanding of 'Super Mario' gradually diminished, making it difficult to consistently produce a cartoon character. Moreover, the layout of the images was also influenced by the random seed. In some instances, even with added descriptions, it was challenging to obtain images that completely matched the target image, as illustrated in the second topic in the Figure \ref{failure}.

\end{document}